\documentclass{article}
\usepackage{MnSymbol} 
\usepackage{caption}
 \usepackage{algorithm}
\usepackage{algorithmic}
\usepackage{graphics}
\usepackage{graphicx}
\usepackage{multirow}
\usepackage[font=small]{caption}
\usepackage{array}
\usepackage{enumitem}
\usepackage{microtype}
\usepackage{graphicx}
\usepackage{subfigure}
\usepackage{booktabs} 

\usepackage{hyperref}

\usepackage[accepted]{icml2023}

\usepackage{amsmath}
\usepackage{mathtools}
\usepackage{amsthm}

\usepackage[capitalize,noabbrev]{cleveref}

\theoremstyle{plain}
\newtheorem{theorem}{Theorem}[section]

\newtheorem{lemma}[theorem]{Lemma}

\theoremstyle{definition}
\newtheorem{definition}[theorem]{Definition}

\theoremstyle{remark}
\newtheorem{remark}[theorem]{Remark}

\usepackage[textsize=tiny]{todonotes}

   \usepackage{amsfonts}
   \usepackage{mathrsfs}

  \newcommand{\EE}{\mathbb{E}}

\icmltitlerunning{Theoretical Guarantees of Learning Ensembling Strategies with Applications to Time Series Forecasting}

\begin{document}

\twocolumn[
\icmltitle{Theoretical Guarantees of Learning Ensembling Strategies \\ with Applications to Time Series Forecasting}

\begin{icmlauthorlist}
\icmlauthor{Hilaf Hasson}{comp}
\icmlauthor{Danielle C. Maddix}{comp}
\icmlauthor{Yuyang Wang}{comp}
\icmlauthor{Gaurav Gupta}{comp}
\icmlauthor{Youngsuk Park}{comp}
\end{icmlauthorlist}

\icmlaffiliation{comp}{AWS AI Labs, Santa Clara, CA, USA}

\icmlcorrespondingauthor{Hilaf Hasson}{hashilaf@amazon.com}

\icmlkeywords{Machine Learning, ICML}

\vskip 0.3in
]

\printAffiliationsAndNotice{} 

\begin{abstract}
  Ensembling is among the most popular tools in machine learning (ML) due to its effectiveness in  minimizing variance and thus improving generalization. Most ensembling methods for black-box base learners fall under the umbrella of ``stacked generalization,'' namely training an ML algorithm that takes the inferences from the base learners as input. While stacking has been widely applied in practice, its theoretical properties are poorly understood. In this paper, we prove a novel result, showing that choosing the best stacked generalization from a (finite or finite-dimensional) family of stacked generalizations based on cross-validated performance does not perform ``much worse'' than the oracle best. Our result strengthens and significantly extends the results in \citet{van2007super}.  Inspired by the theoretical analysis, we further propose a particular family of stacked generalizations in the context of probabilistic forecasting, each one with a different sensitivity for how much the ensemble weights are allowed to vary across items, timestamps in the forecast horizon, and quantiles. Experimental results demonstrate the performance gain of the proposed method.  
 
\end{abstract}
\section{Introduction}
Ensemble methods have been a staple in machine learning with ensemble-based methods such as Random Forests \cite{breiman2001random} and XGBoost \cite{chen2016xgboost} being among the most popular choices for tabular data by practitioners \cite{KS2020, KS2021}. 
Both classical (mean, median, weights based on cross-validated performance) and relatively modern ensembling strategies, e.g., \citet{erickson2020autogluon} (which in two popular Kaggle competitions has beat $99\%$ of the participating data scientists after training on the raw data) are examples of ``stacked generalizations'', or ``stacking'' \cite{wolpert1992stacked}; see \cite{ting1997stacking}. Stacking is an umbrella term referring to the process of training one model on top of the predictions of the base learners. In the simple mean and median cases, the stacking model is constant rather than learned. Despite being heavily used in practice, the theoretical properties of stacked generalizations are underexplored, and stacking has notoriously been referred to as ``black art'' \cite{wolpert1992stacked, ting1997stacked}.

One early attempt to quantifiably understand stacked generalization is due to \citet{van2007super}. Built on top of \citet{van2006oracle}, the authors prove a theoretical guarantee (Theorem 2 of \citet{van2007super}) that choosing the best stacked generalization out of a \emph{finite} (rather than \emph{finite-dimensional}) family of \emph{constant} (as opposed to \emph{learned}) stacked generalizations based on cross-validation performance does not do much worse than the oracle best. They then apply their result to the case of a single learned stacked generalization by discretizing the space of the functions that the stacked generalization may become after the training, resulting in a guarantee that degrades with the cardinality of the chosen discretization.

In this paper, we make a major step towards a better theoretical understanding of the stacking mechanism. More precisely, we show new results that extend \citet{van2007super} in two ways: 1. We remove the assumption that the stacked generalizations are constant, and permit them to be learned, without changing the conclusion of the result. This allows us to get much better guarantees even for the case of a single stacked generalization, as it removes the need for a discretization. (See the discussion following Theorem \ref{mainstack}.) 2. We extend the result from the case of finite families to the case of finite-dimensional families.

We further push the boundary beyond the tabular case to study the concrete use case of time series forecasting. Ensembling of time series models is a notoriously challenging task, coined as the ``Forecast Combination Puzzle'' by \cite{stock2004combination}, with many follow-up works in the literature; see Section \ref{related}. We concern ourselves with datasets that involve multiple time series, forecasting a fixed number of timestamps into the future, and predicting multiple quantiles for each time series and timestamp. Motivated by the theoretical insights, we propose a particular finite-dimensional family of stacked generalizations, in which how much the ensemble weights are allowed to vary across time series, timestamps, and quantiles is each controlled by a dimension of the family.

In summary, our contributions are two-fold:
\begin{enumerate}

    \item In the tabular case we show (Theorem \ref{mainstack}) that the process of letting cross validation determine a stacked generalization from an arbitrary finite or finite-dimensional family of stacked generalizations cannot perform much worse than the oracle best, and under an additional assumption, from each of the base learners. This extends a result from \citet{van2007super} by allowing the stacked generalizations to be learned rather than constant, and by extending from the case of a finite to finite dimensional family. In the case of a single stacked generalization, our result gives tighter bounds than \citet{van2007super}, as the latter requires a discretization of the family of potential models of the stacked generalization once trained.
    \item 
    We propose a family of stacked generalizations in the time series use case, as well as a setup for choosing the best performing one, in a manner that learns the ensemble elasticity (how much the weights are allowed to vary) across time series/timestamps/quantiles. Despite Theorem \ref{mainstack} not applying directly to the non-tabular case, we show experiments that demonstrate that our method inspired by this theory 
    is effective, supported by strong evidence that it is able to adjust to changes in performance of the base learners across time series/timestamps/quantiles.

\end{enumerate}

We organize the rest of the paper as follows. Related work is discussed in Section~\ref{related}, followed by preliminaries and notation (Section~\ref{prelim}) to set the stage. The subsequent sections (\ref{sec:main} and \ref{sec:theorem}) cover the main results with its proof. In Section~\ref{timefamily}, we present the case study of time series forecasting, and use the next two sections to provide empirical evidences for the effectiveness of the proposed algorithm before concluding the paper.

\section{Related Work}\label{related}
In this section, we give an overview of theoretical and applied research on stacked generalizations, as well as their use in time series forecasting.

\paragraph{Stacked Generalizations}
Stacked generalization has become an exceedingly popular choice for ensembling in practice, and particularly impressive in its results, as illustrated in \citet{erickson2020autogluon}. In a recent work \cite{kim2021deep}, it has also been used for ensembling probabilistic predictions in the tabular setup. We refer to \citet{zhou2012ensemble} for a general discussion on the benefits of ensembling, and to \citet{cruz2018dynamic} for a thorough benchmarking of ensembling methods.

We remark that while our paper concerns black box base learners, there has also been an active research front on creating top performing algorithms by choosing the base learners as well as the ensembling method, e.g., \citet{breiman2001random, chen2016xgboost, fort2019deep, balaji2017}, etc. 
\paragraph{Theoretical Results for Stacked Generalizations}
As mentioned in the introduction, in \citet{van2007super} the authors provide theoretical guarantees for stacked generalizations. They first discretize the set of possible functions that the stacked generalization may become once trained, and in this way reduce the problem to giving guarantees (Theorem 2 of \citet{van2007super}) that choosing the best performing (in cross validation) stacked generalization from a finite set of stacked generalizations, each of which is a constant function (e.g., simple mean, or some other constant weighted combination), does not do much worse than the oracle best.

\paragraph{Time Series Ensembling}
In the time series use case, it has been empirically observed that often simple averaging of the forecasts of the base learners is superior to more sophisticated ensemble methods; a problem dubbed the ``forecast combination puzzle'' in \citet{stock2004combination}. This dates back to \citet{bates1969combination}, where the authors observed that attempting to learn covariance terms between the errors of the individual learners introduces too much variance. Similar concerns over the amount of variance introduced when learning weights for time series ensembling are voiced in \citet{smith2009simple} and \citet{claeskens2016forecast}, where both works explored squared loss. \citet{elliott2011averaging} focuses on understanding the potential gain from learning the optimal weights compared to simple averaging, and provides the following two results: first, that if the maximal eigenvalue of the covariance matrix of errors of the base learners is bounded, then as the number of base learners goes to infinity, the performance of using the optimal weights converges to that of using simple average; second, he gives an upper bound for the potential gain for the case of $3$ and $4$ base learners. We remark that the papers referenced here regarding the forecast combination puzzle are restricted to the situation of learning weights that are global (unchanging across time or any other dimension), and based on an analysis of mean squared loss (so as to simplify the theory), and they do not apply more broadly.

Stacked generalizations have been applied for time series forecast ensembling as far back as in \citet{donaldson1996forecast, moon2020combination, massaoudi2021novel}. In \citet{gastinger2021study} we see the success of these single stacked generalizations in a large empirical study. In all of these cases a single stacked generalization was used, and not a family of stacked generalizations.

\section{Preliminaries and Notation}\label{prelim}

In both the main theorem and in its proof we use language and notation commonly used in oracle inequalities. In order to be self contained we will now introduce the relevant notation and terminology. 

Let $\{\theta_{\alpha}\}_{\alpha \in \mathcal{J}}$ be a finite set of algorithms for tabular data. Assume that feature space is $d$ dimensional, the target values are $r$ dimensional, and the output of predictions is $r'$ dimensional. (An example when $r\neq r'$ would be: $r=1$, and the predictions are for $r'$ many conditional quantiles.) We write $\theta_{\alpha}(D_0):\mathbb{R}^d\rightarrow\mathbb{R}^{r'}$ for the model that results from letting $\theta_{\alpha}$ train on a training dataset $D_0$. (For the sake of simplicity, assume that $\theta_{\alpha}(D_0)$ is completely determined by $D_0$, though this assumption can easily be removed.) Let $L:\mathbb{R}^{r'}\times\mathbb{R}^r\rightarrow\mathbb{R}$ be a fixed loss function, and let $L_{\alpha}(D_0):\mathbb{R}^d\times\mathbb{R}^r\rightarrow\mathbb{R}$ be defined by $L_{\alpha}(D_0)((x,y)):=L(\theta_{\alpha}(D_0)(x),y)$.

Let $X$ be a random sample of $\mathbb{R}^d\times\mathbb{R}^r$ w.r.t. the same probability distribution $P$ as the one being sampled by the data.\footnote{This notation is common in literature, but note that it is confusing: $X$ is both the feature vector and the target.} The $\tilde{\alpha}$ that minimizes $\EE(\int L_{\tilde \alpha}(D_0)(X)dP)$ (where the expected value runs over $D_0$) is called the ``oracle''. We never have direct access to the oracle, but given a validation set $D_1$ we can choose the $\hat{\alpha}$ for which $\theta_{\hat{\alpha}}$ has the lowest (empirical) validation loss. In order to be more precise, write $D_i=\{X_1^i,...,X_{n_i}^i\}$ for $i=0,1$ (where the $\{X_l^i\}_{l=1}^{n_i}$ are IID random variables taking values in $\mathbb{R}^d\times\mathbb{R}^r$ for some fixed $n_i\in\mathbb{N}$), representing training ($i=0$) and validation $(i=1)$ sets; and let $\mathbb{P}_i$ for $i=0,1$ be the empirical distribution $\frac1{n_i}\sum_{l=1}^{n_i} \delta_{X_l^i}$. Then $\hat{\alpha}$ is an index for which $\int L_{\hat \alpha}(D_0)(X)d\mathbb{P}_1$ is minimized over $\alpha\in\mathcal{J}$.

Oracle inequalities are about giving upper bounds of $\EE(\int L_{\hat \alpha}(D_0)(X)dP)$ in terms of $\EE(\int L_{\tilde \alpha}(D_0)(X)dP)$, quantifying how much worse $\hat{\alpha}$ is compared with the oracle. (While much of the relevant literature accommodates a scheme where the data is split multiple times to train and validation, we have chosen to focus on the case of a single split for the sake of simplified notation. The interested reader should be able to easily generalize.) In Section \ref{sec:theorem} we introduce a new oracle inequality that allows us to prove our main theorem (Theorem \ref{mainstack}).

In the oracle inequalities we discuss we will use the following notion.
\begin{definition}
Let $\mathcal{X}$ be a sample space with probability measure $P$. Then a we say that a function $f:\mathcal{X}\rightarrow\mathbb{R}$ has Bernstein numbers (or Bernstein pair) $(M(f), v(f))\in\mathbb{R}^2$ if:
\begin{equation}
  M(f)^2\int\left(e^{\frac{|f|}{M(f)}}-1-\frac{|f|}{M(f)}\right)dP\leq \frac{1}{2}v(f).
\end{equation}
A set of functions $\mathcal{F}$ from $\mathcal{X}$ to $\mathbb{R}$ is said to have Bernstein numbers $(M(\mathcal{F}), v(\mathcal{F}))\in\mathbb{R}^2$ if they are Bernstein numbers for all $f\in\mathcal{F}$.
\end{definition}
The existence of Bernstein numbers can be viewed as a weak moment condition. (It is a much weaker condition than the functions in $\mathcal{F}$ having a uniformly bounded range; we refer to \cite{wellner2013weak} for further intuition.) 

Finally, for any subset $\mathcal{J}$ of Euclidean space and value $\varepsilon>0$ the notation $N^{\text{int}}(\mathcal{J},\varepsilon)$ denotes the (internal) covering number of the space $\mathcal{J}$ with respect to its ambient Euclidean space.

\section{Main Theorem}
\label{sec:main}

Let $D_0$ and $D_1$ be datasets of, respectively, $n_0$ and $n_1$ IID random variables, each single variable taking values in $\mathbb{R}^d\times\mathbb{R}^r$ (i.e, with feature dimension $d$ and target dimension $r$).  

Let $\eta_1,...,\eta_m$ be a finite set of tabular algorithms (henceforth ``the base learners'') with feature dimension $d$, prediction dimension $r'$, and target dimension $r$. Let $\{A_{\alpha}\}_{\alpha\in\mathcal{J}}$ be a set of stacked generalizations (tabular algorithms with feature dimension $m$, 
prediction dimension $r'$, and target dimension $r$). Let $D_0$ be further split as $D_0=D_{00}\cupdot D_{01}$. The base learners will train on $D_{00}$, whereas the stacked generalizations will train on the predictions of the base learners on $D_{01}$. (We remark that having $D_{00}$ and $D_{01}$ be disjoint is the method recommended in \cite{zhou2012ensemble}; though in fact what follows will continue to hold if you set $D_0=D_{00}=D_{01}$.)

Let
\[Z:=\{((\eta_1(D_{00})(x),...,\eta_m(D_{00})(x)), y)|(x,y)\in D_{01}\}\]
be the predictions on $D_{01}$ of the base learners that were trained on $D_{00}$. Let $L:\mathbb{R}^{r'}\times\mathbb{R}^r\rightarrow\mathbb{R}$ be some loss function, and let 
$L_{\alpha}(D_{00}, D_{01})((x,y)):=L(A_{\alpha}(Z)(\eta_1(D_{00})(x),...,\eta_m(D_{00})(x)),y),$
where $A_{\alpha}(Z)$ stands for $A_{\alpha}$ trained on $Z$.

Consider $\mathcal{F}:=\{(x,y)\mapsto L_{\alpha}(D_{00}, D_{01})((x,y))|\alpha \in \mathcal{J}\}$, and let $(M(\mathcal{F}), v(\mathcal{F}))$ be its Bernstein numbers. Finally, let $X$ be a random variable in $\mathbb{R}^d\times\mathbb{R}^r$ following the same distribution that $D_0$ and $D_1$ are sampling from. Then the following holds, using the notation in Section \ref{prelim}.
\begin{theorem}\label{mainstack}
With the notation above, for every
$ \hat{\alpha}\in \mathcal{J}$ let
\begin{equation*}
\begin{aligned}
  W_{\hat{\alpha}}&:=\Big\{\tilde{\alpha}\in \mathcal{J}\Big|
  \int L_{\hat \alpha}(D_{00}, D_{01})(X)d\mathbb{P}_1\\
  &\leq \int L_{\tilde \alpha}(D_{00}, D_{01})(X)d\mathbb{P}_1\Big\},  
  \end{aligned}
\end{equation*}
be the set of indices $\tilde{\alpha}$ for which $\theta_{\hat{\alpha}}$ outperforms $\theta_{\tilde{\alpha}}$ on validation. Further assume that $\mathcal{J}$ is a bounded subset of a finite dimensional Euclidean space, and that for every $D_0$ the function that takes $\alpha \in \mathcal{J}$ to $L_{\alpha}(D_{00}, D_{01})\in\mathcal{F}$ is Lipschitz with constant $\ell$ w.r.t. the infinity norm in $\mathcal{F}$. Then we have that for every $\delta>0$, $1\leq p\leq 2$, and sequence $\varepsilon_{n_1}>0$,
\begin{align}\label{eq:result}
\begin{aligned}
    &\EE(\int L_{\hat \alpha}(D_{00}, D_{01})(X)dP) \leq\\& (1+2\delta)\inf_{\tilde{\alpha}\in W_{\hat{\alpha}}}\left(\EE(\int L_{\tilde \alpha}(D_{00}, D_{01})(X)dP)\right) \\
    &\quad{+}\:\sup_{f\in\mathcal{F}}(B_f)+2\Big((1+\delta)+\frac{1}{n_1}\Big)\varepsilon_{n_1},\\
\end{aligned}
\end{align}
where
\[
    B_f:=\frac{16\left(\frac{M(\mathcal{F})}{n_1^{1-\frac1p}} + \left(\frac{v(\mathcal{F})}{(\delta\int fdP)^{2-p}}\right)^{\frac1p}\right)\log(1+N^{\text{int}}(\mathcal{J},\frac{\varepsilon_{n_1}}{\ell}))}{n_1^{\frac1p}}.
\]
In addition, the following hold:
\begin{enumerate}[noitemsep,topsep=0pt] 
\item If we choose $\varepsilon_{n_1}:=n_1^{-\frac12-\epsilon}$ for any $\epsilon>0$ then $2\Big((1+\delta)+\frac{1}{n_1}\Big)\varepsilon_{n_1}=O(n_1^{-\frac12-\epsilon})$ (in particular, $o(1)$). Moreover, since $\mathcal{J}$ is bounded in finite-dimensional Euclidean space, $\log(1+N^{\text{int}}(\mathcal{J},\frac{\varepsilon_{n_1}}{\ell}))=O(\log n_1)$. For $p=2$ (so that $(\delta\int f dP)^{2-p}=1$), we consequently have $\sup_{f\in\mathcal{F}}(B_f)=O(\log n_1/\sqrt{n_1})=o(1)$. For $p<2$, the same conclusion holds provided $\inf_{f\in\mathcal{F}}\int f dP \ge c$ for some constant $c>0$ (so that $\sup_{f\in\mathcal{F}}(B_f)=O(\log n_1/n_1^{\frac1p})=o(1)$).
    \item For the guarantee to hold not only against the best stacked generalizations in the family, but also against the original base learners, one simply adds the summand $m$ to the inside of the $\log$ term.
    \item If $\mathcal{J}$ is finite rather than finite dimensional, then the log term can be replaced with $\log(1+|\mathcal{J}|)$, and we may remove the last summand $2\Big((1+\delta)+\frac{1}{n_1}\Big)\varepsilon_{n_1}$.

\end{enumerate}

\end{theorem}

In the case that $|\mathcal{J}|$ is finite, and that, in addition, the stacked generalizations are constant functions (i.e., do not depend on $Z$ at all), then the theorem above specializes to Theorem 2 of \citet{van2007super}. The two main differences are the following: 
\begin{enumerate}
[noitemsep,topsep=0pt] 
    \item There the authors make additional assumptions (such as that $r=r'=1$ and that the loss function is $MSE$); and they write in the language of multiple cross validations, which we have chosen not to use so as not to overload notation (though generalization is straightforward).
    \item In \citet{van2007super}, the authors argue that their result applies to the case that $|\mathcal{J}|=1$ and the stacked generalization is learned (does depend on $Z$) by first discretizing the set of functions that the stacked generalization may become, thus reducing to the case of a finite number of constant stacked generalizations $\tilde{\mathcal{J}}$ as in Theorem 2 therein; and that the error term from this step is asymptotically negligible. Note, however, that by following this procedure the log term becomes $\log(1+|\tilde{\mathcal{J}}|+m)$ (which explodes with the size of the discretization); whereas if you use the theorem above you get $\log(2+m)$ and does depend on a discretization. Therefore the theorem above is a significant improvement even in the case $|\mathcal{J}|=1$. 
\end{enumerate}

\section{Reduction to Oracle Inequalities}\label{sec:theorem}
One of the main insights that led to Theorem \ref{mainstack} is the recognition that a sufficiently general and strong version of an oracle inequality (Theorem \ref{mainthm} below) can lead to significant gains in obtaining better and more general theoretical guarantees for stacked generalizations, as we proceed to illustrate.

We deviate from common notation by now letting $D_0$ and $D_1$ be \it ordered \rm datasets of IID feature and target pairs; taking values in $(\mathbb{R}^d\times\mathbb{R}^r)^{n_0}$ and $(\mathbb{R}^d\times\mathbb{R}^r)^{n_1}$ respectively. Let $X$ be a random variable in $\mathbb{R}^d\times\mathbb{R}^r$ following the same distribution. Let $\{\theta_{\alpha}\}_{\alpha\in\mathcal{J}}$ be a set of tabular algorithms that may take into account the order of the training dataset. Let $L:\mathbb{R}^{r'}\times\mathbb{R}^r\rightarrow\mathbb{R}$ be some loss function, and let $L_{\alpha}(D_0)((x,y)):=L(\theta_{\alpha}(D_0)(x),y)$. Finally let $\mathcal{F}:=\{(x,y)\mapsto L_{\alpha}(D_0)((x,y))|\alpha \in \mathcal{J}\}$, and let $(M(\mathcal{F}), v(\mathcal{F}))$ be its Bernstein numbers. Then the following holds.  

\begin{theorem}\label{mainthm}
With the notation above, for every
$ \hat{\alpha}\in \mathcal{J}$ let
\begin{equation*}
\begin{aligned}
  W_{\hat{\alpha}}&:=\Big\{\tilde{\alpha}\in \mathcal{J}\Big|\\&
  \int L_{\hat \alpha}(D_0)(X)d\mathbb{P}_1\leq \int L_{\tilde \alpha}(D_0)(X)d\mathbb{P}_1\Big\},  
  \end{aligned}
\end{equation*}
be the set of indices $\tilde{\alpha}$ for which $\theta_{\hat{\alpha}}$ outperforms $\theta_{\tilde{\alpha}}$ on validation. Further assume that $\mathcal{J}$ is a bounded subset of a finite dimensional Euclidean space, and that for every $D_0\in(\mathbb{R}^d\times\mathbb{R}^r)^{n_0}$ the function that takes $\alpha \in \mathcal{J}$ to $L_{\alpha}(D_0)\in\mathcal{F}$ is Lipschitz with constant $\ell$ w.r.t. the infinity norm in $\mathcal{F}$. Then we have that for every $\delta>0$, $1\leq p\leq 2$, and sequence $\varepsilon_{n_1}>0$,
\begin{align}\label{eq:resulttwo}
\begin{aligned}
    &\EE(\int L_{\hat \alpha}(D_0)(X)dP) \leq\\& (1+2\delta)\inf_{\tilde{\alpha}\in W_{\hat{\alpha}}}\left(\EE(\int L_{\tilde \alpha}(D_0)(X)dP)\right) \\
    &\quad{+}\:\sup_{f\in\mathcal{F}}\left(B_f\right)+2\Big((1+\delta)+\frac{1}{n_1}\Big)\varepsilon_{n_1},\\
\end{aligned}
\end{align}
where
\[
    B_f:=\frac{16\left(\frac{M(\mathcal{F})}{n_1^{1-\frac1p}} + \left(\frac{v(\mathcal{F})}{(\delta\int fdP)^{2-p}}\right)^{\frac1p}\right)\log(1+N^{\text{int}}(\mathcal{J},\frac{\varepsilon_{n_1}}{\ell}))}{n_1^{\frac1p}}.
\]
In addition, the following hold:
\begin{enumerate}
[noitemsep,topsep=0pt] 
\item If we choose $\varepsilon_{n_1}:=n_1^{-\frac12-\epsilon}$ for any $\epsilon>0$ then $2\Big((1+\delta)+\frac{1}{n_1}\Big)\varepsilon_{n_1}=O(n_1^{-\frac12-\epsilon})$ (in particular, $o(1)$). Moreover, since $\mathcal{J}$ is bounded in finite-dimensional Euclidean space, $\log(1+N^{\text{int}}(\mathcal{J},\frac{\varepsilon_{n_1}}{\ell}))=O(\log n_1)$. For $p=2$ (so that $(\delta\int f dP)^{2-p}=1$), we consequently have $\sup_{f\in\mathcal{F}}(B_f)=O(\log n_1/\sqrt{n_1})=o(1)$. For $p<2$, the same conclusion holds provided $\inf_{f\in\mathcal{F}}\int f dP \ge c$ for some constant $c>0$ (so that $\sup_{f\in\mathcal{F}}(B_f)=O(\log n_1/n_1^{\frac1p})=o(1)$).
    \item If $\mathcal{J}$ is finite rather than finite dimensional, then the log term can be replaced with $\log(1+|\mathcal{J}|)$, and we may remove the last summand $2\Big((1+\delta)+\frac{1}{n_1}\Big)\varepsilon_{n_1}$.
    \item If one wants to artificially add to $W_{\hat \alpha}$ any finite number $p$ of additional algorithms, then, in both the case that $\mathcal{J}$ is finite and finite-dimensional, one simply adds $p$ to the inside of the $\log$ term. 
\end{enumerate}

\end{theorem}
\begin{remark}\,
    We remark that $\ell$ can easily be discerned from the Lipschitz constant of $L$ in its first coordinate and the Lipschitz constants of each of the maps $\alpha\mapsto \theta_{\alpha}$. Note also that if $\mathcal{J}$ is compact and $\alpha\mapsto \int L_{\alpha}(D_0)(X)d\mathbb{P}_1$ is continuous for each fixed $(D_0,D_1)$, then there exists a (measurable) minimizer $\hat{\alpha}$ of the empirical validation loss, in which case $W_{\hat{\alpha}}=\mathcal{J}$. In this case $\hat{\alpha}$ has the interpretation of being the index of the best performing stacked generalization on cross validation, and $\inf_{\tilde{\alpha}\in \mathcal{J}}\left(\EE(\int L_{\tilde \alpha}(D_0)(X)dP)\right)$ is achieved by some $\tilde{\alpha}$ that has the interpretation of being the oracle best stacked generalization. 
\end{remark}

We postpone the proof of this theorem to Appendix \ref{omitted}, though it is important to point out that this is both a strengthening and a generalization of Lemma 2.3 (or more accurately of Lemmas 2.1 and 2.2) of \cite{van2006oracle} to the case where the datasets are ordered and to the case that $\mathcal{J}$ is potentially infinite. This is precisely the version that we require to prove Theorem \ref{mainstack} easily.
\begin{proof} (of Theorem \ref{mainstack})\,\\
Apply Theorem \ref{mainthm} to the set of algorithms $\theta_{\alpha}(D_0):=A_{\alpha}(Z)\circ(\eta_1(D_{00}),...,\eta_m(D_{00}))$.
\end{proof}

Note that even in the case that $\mathcal{J}$ is finite it would not have been possible to use the oracle inequalities in \citet{van2006oracle} directly to the algorithms $\theta_{\alpha}(D_0):=A_{\alpha}(Z)\circ(\eta_1(D_{00}),...,\eta_m(D_{00}))$ since these $\theta_{\alpha}$'s depend on the order of $D_0$ (to separate out $D_{00}$ and $D_{01}$); and that this is the reason that \citet{van2007super} first reduced to the case of constant stacked generalizations (via a discretization of the set of functions that the stacked generalization may become).

\section{Case Study: Probabilistic Time Series Forecasting}\label{timefamily}
Consider probabilistic forecasting algorithms that output a prediction for each time series in the dataset (henceforth ``item''), for each timestamp in the forecast horizon, and for each of a list of predefined quantiles. In combining these algorithms into a weighted sum, it is a natural question whether it is a good idea to share the weights across items/timestamps/quantiles. As we will see, this translates naturally to choosing the best stacked generalization out of a family. While Theorem \ref{mainstack} does not apply directly outside of the tabular use case, we use it as motivation.

\subsection{Setup for Stacked Generalizations in Forecasting}
As is common in probabilistic time series forecasting, we consider a dataset containing $N$ time series (henceforth ``items''). We consider a base learners $\eta_1,...,\eta_m$, where each one of them makes predictions for a fixed number of timestamps in the future $h$ (henceforth ``the forecast horizon''), and the output for each item and timestamp is a prediction of some fixed predicted quantiles $q$ (e.g., $q=3$ with quantiles $0.1, 0.5, 0.9$). In particular the output for each base learner is in $\mathbb{R}^{N\times h\times q}$. (See Appendix \ref{sec:pts_prob_def} for details.) 

Our interpretation of a stacked generalization in this setup is an algorithm that takes as input the output of all of the base learners trained up to some backtest window together with the true values from this backtest window, and outputs for each tuple (item, timestamp in the forecast horizon, and quantile) a weight for each of the base learners. (One example of such a stacked generalization is one that optimizes mean weighted quantile loss over the backtest window treating all of the weights as independent; another example would be one that optimizes mean weighted quantile loss on the backtest window, but with each base learner having the same weight regardless of the timestamp, item, and quantile. We refer to Appendix \ref{sec:pts_prob_def} for the definition of mean weighted quantile loss.) 

Once the weights are learned in this fashion on a single backtest window, the inference of the stacked generalization is a weighted sum of the base learners, where the weights vary based on item, timestamp, and quantile. 

\begin{figure}[h]
	\centering
	\includegraphics[width=.45\textwidth]{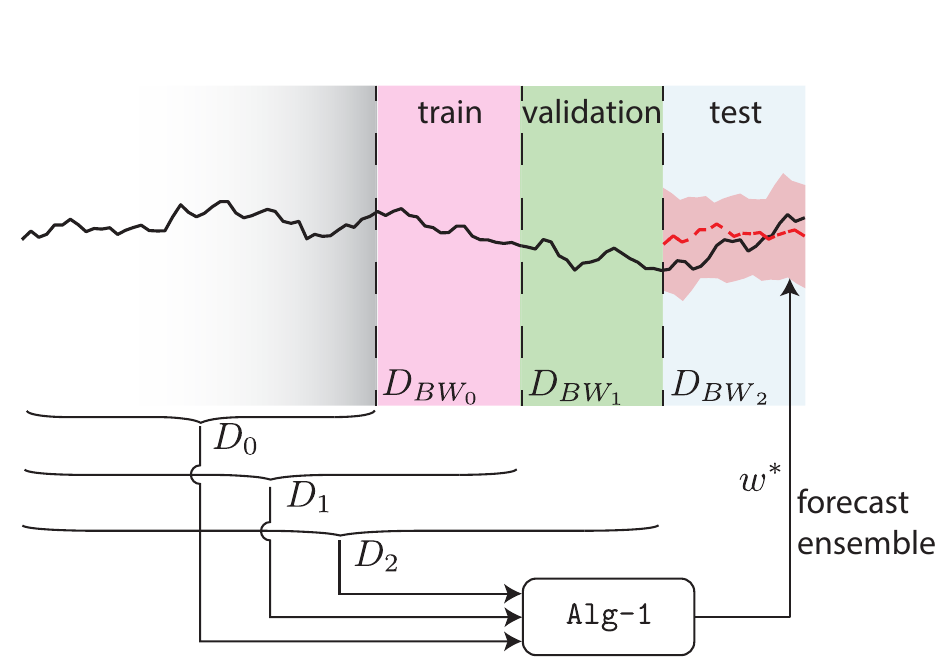}
	\caption{Dataset split for the purpose of choosing the best stacked generalization out of a family based on cross validatation. (See Algorithm \ref{alg:compute_weights}.) Each $D_{\text{BW}_i}$ contains as many timestamps as the forecast horizon length of the base learners. The split is done across all items.}
	\label{fig:bw_process}
\end{figure}

Given a family of stacked generalizations, we interpret choosing the best performing one based on cross validation as specified in Algorithm \ref{alg:compute_weights}, where the notation is borrowed from Figure \ref{fig:bw_process}.

\label{sec:two_step_algo}

\begin{algorithm}[tb]
    \caption{Choosing a Stacked Generalization Out of Family in the Time Series Setting}\label{alg:example}
    \begin{algorithmic}[1]
    \renewcommand{\algorithmicrequire}{\textbf{Input:}}
    \REQUIRE $m$ base learners: $\eta_1, \dots, \eta_m$, a time series dataset split as in Figure \ref{fig:bw_process}; and a family of stacked generalizations $\{A_{\alpha}\}_{\alpha\in\mathcal{J}}$.\\
    \renewcommand{\algorithmicensure}{\textbf{Output:}}
    \ENSURE Predictions for $D_{\text{BW}_2}$.\\
    \FOR{values of $\alpha$}
    \STATE Train $\eta_1,...,\eta_m$ on $D_0$ and use their predictions together with $D_{\text{BW}_0}$ to train $A_{\alpha}$. 
    Store the resulting weights $w$.
    \STATE Train $\eta_1,...,\eta_m$ on $D_1$ and $w$ to make a weighted sum prediction on $D_{\text{BW}_1}$; compute mean weighted quantile loss.
    \ENDFOR
    \STATE Store the $\hat{\alpha}$ for which the mean weighted quantile loss computed was minimal. 
    \STATE Retrain $A_{\hat{\alpha}}$ on the output of the models $\eta_1,...,\eta_m$ trained on $D_1$, together with $D_{\text{BW}_1}$. Store the resulting weights as $w^*$.
    \STATE Train $\eta_1,...,\eta_m$ on $D_2$ and output their weighted sum using $w^*$.
\end{algorithmic}
\label{alg:compute_weights}
\end{algorithm}

We remark that a uniform notation between tabular and the time series use cases it possible, albeit confusing. We refer to Appendix \ref{samenotation} for such a treatment, while in the remainder of the paper we make use of simplified notation.

\subsection{A Family of Stacked Generalizations Controlling Uniformity of Weights}\label{sec:givefamily}
We propose a family of stacked generalizations indexed by $\alpha\in\mathbb{R}_{\geq 0}^4$; taking the output of the base learners together with true values for the backtest window on which they are making predictions, and finding the weights that optimize mean weighted quantile loss (see Appendix \ref{sec:pts_prob_def} for the definition) together with the regularizers:
\[
\sum_{d=1}^3\alpha_d\mathbb{H}(\sigma^{(d)}(w))
+ \alpha_{4}\sum\limits_{i,j,k,l}|w^{(l)}_{i,j,k}|,
\]
where $\mathbb{H}(\sigma^{(d)}(w))$ is an entropy term defined in detail in Appendix \ref{app:method} that controls whether the weights are forced to be uniform across items, timestamps, and quantiles. Thus, for example, the algorithm $A_{(0,0,0,0)}$ optimizes mean weighted quantile loss treated all the weights as independent; whereas $A_{(\alpha_1,0,0,0)}$ as $\alpha_1$ goes to infinity would optimize mean weighted quantile loss under the constraint that for a specific timestamp and quantile the base learner weights don't vary across items.

\subsection{Experiments}\label{experiments}
In this subsection, we train base learners $\{\eta_l(D_2)\}_l$ on the full historical data $D_2$, and compare our optimal stacked generalization ensemble $A_{\hat{\alpha}}(D_2)$ in Eqn. \eqref{eqn:final_ensemble} to the predictions from the base learners and simple ensembling methods. We report the mean weighted quantile loss (Eqn. \eqref{eq:mean_QL}) over the quantiles $\tau = [0.1, 0.5, 0.9]$ on the test set $L( A_{\hat{\alpha}}(D_2), D_{\text{BW}_2})$  on the corresponding most recent in time backtest window $D_{\text{BW}_2}$ (See Eqn. \eqref{eq:train_val} to recall the definition of $(D_2, D_{\text{BW}_2})$).  For each $\alpha$, we use autodiff in PyTorch \cite{Paszke17diff} to find the optimal weights for each $A_{\alpha}$ in Algorithm \ref{alg:compute_weights} (using softmax to satisfy the constraint), and use scipy's implementation of COBYLA \cite{powell1994direct} to find the optimal $\hat \alpha$ since looping over an infinite set is impossible. (It is possible to do a grid search to find $\hat\alpha$, but we have empirically observed that using an optimization method is faster and leads to equivalent results.) 

We test on real-world open-source datasets from the UCI data repository \cite{Dua:2017}, Kaggle \cite{lai_dataset_2017}, and \texttt{M4} competition datasets \cite{makridakisM4concl} (see Table \ref{tab:datasets} in Appendix \ref{detailsappendix}).

\paragraph{Base Learners.}
We call the following $m=6$ base learners from GluonTS \cite{gluonts_jmlr} with the default hyper-parameter settings:  the local state space models ARIMA and ETS \footnote{The error from the ETS predictions on \texttt{Wiki} is too large to include, and so we omit ETS in the corresponding ensembles.} \cite{hyndman2008forecasting}, Non-Parametric Time Series (NPTS) \cite{gasthaus2016}, the global deep learning models DeepAR \cite{flunkert2017deepar} and MQ-CNN \cite{wen2017multi}, and Rotbaum (XGBoost-based forecaster, a GluonTS implementation of TTSW from \citet{hasson2021probabilistic}. We compare to the simple ensemble baselines in Appendix \ref{sec:baselines}. The experiments are performed $5$ times to compute confidence intervals.
\begin{table*}[h]
\centering
\caption{Mean weighted quantile losses over quantiles $\tau = [0.1, 0.5, 0.9]$ for various base learners and ensembling strategies on real-world open-source time series datasets over 5 runs.}
\label{tab:regular}
\vspace{0.1in}
\resizebox{0.75\linewidth}{!}{
\begin{tabular}{cccccc}
\toprule
Base Learner / \\Ensemble Strategy & \texttt{Elec} &\texttt{Kaggle}&\texttt{M4-daily} &\texttt{Traf} &\texttt{Wiki}\\ 
\cmidrule(l){1-6}
ARIMA&0.105 $\pm$ 0.0&0.1201 $\pm$ 0.0&0.0341 $\pm$ 0.0&0.2219 $\pm$ 0.0&0.6008 $\pm$ 0.0\\
DeepAR&0.0585 $\pm$ 0.0011&0.2478 $\pm$ 0.0056&0.035 $\pm$ 0.0009&0.0987 $\pm$ 0.0014&0.3311 $\pm$ 0.0095\\
ETS&0.0796 $\pm$ 0.0203&\textbf{0.1193 $\pm$ 0.0004}&0.033 $\pm$ 0.0005&0.2999 $\pm$ 0.0&N/A\\
MQ-CNN&0.0698 $\pm$ 0.0018&0.2452 $\pm$ 0.0018&0.0408 $\pm$ 0.0025&0.2491 $\pm$ 0.0423&0.3639 $\pm$ 0.0038\\
NPTS&0.0536 $\pm$ 0.0005&0.1393 $\pm$ 0.0001&0.1218 $\pm$ 0.0001&0.1172 $\pm$ 0.0002&0.4798 $\pm$ 0.0001\\
Rotbaum&0.0603 $\pm$ 0.001&0.2098 $\pm$ 0.0&0.0342 $\pm$ 0.0003&0.1289 $\pm$ 0.0004&0.3983 $\pm$ 0.0022\\
\midrule
mean&0.0616 $\pm$ 0.0034&0.1556 $\pm$ 0.0006&0.0431 $\pm$ 0.0002&0.1458 $\pm$ 0.004&0.3866 $\pm$ 0.0029\\
median&0.0549 $\pm$ 0.0012&0.1323 $\pm$ 0.0005&0.0351 $\pm$ 0.0001&0.123 $\pm$ 0.0038&0.3489 $\pm$ 0.0027\\
GB ($D_1, D_{\text{BW}_1}$) &0.0531 $\pm$ 0.0004&0.16 $\pm$ 0.0012&0.0327 $\pm$ 0.0001&0.0987 $\pm$ 0.0014&0.3412 $\pm$ 0.0061\\
Best ($D_1, D_{\text{BW}_1}$) &0.056 $\pm$ 0.0024&0.1201 $\pm$ 0.0&0.0332 $\pm$ 0.0005&0.0987 $\pm$ 0.0013&0.3339 $\pm$ 0.0148\\
Unregularized&\textbf{0.0475 $\pm$ 0.0005}&0.1979 $\pm$ 0.0009&0.028 $\pm$ 0.0001&0.0986 $\pm$ 0.0017&0.3366 $\pm$ 0.0008\\
Ours&0.0494 $\pm$ 0.0006&0.1663 $\pm$ 0.0003&\textbf{0.0266 $\pm$ 0.0003}&\textbf{0.094 $\pm$ 0.001}&\textbf{0.3187 $\pm$ 0.002}\\
\bottomrule
\end{tabular}%
}
\end{table*}

\begin{table*}[h]
\centering
\caption{Mean weighted quantile losses over quantiles $\tau = [0.1, 0.5, 0.9]$ for various base learners and ensembling strategies on real-world open-source time series datasets, where Gaussian noise is added to the base learner predictions across time over 5 runs.}
\label{tab:syn_time}
\vskip 0.1in
\resizebox{0.75\linewidth}{!}{

\begin{tabular}{cccccc}
\toprule
Base Learner / \\ Ensemble Strategy & \texttt{Elec} &\texttt{Kaggle}&\texttt{M4-daily} &\texttt{Traf} &\texttt{Wiki}\\ 
\cmidrule(l){1-6}
ARIMA&1.9146 $\pm$ 0.1236&1.102 $\pm$ 0.0038&0.0367 $\pm$ 0.0001&2.9926 $\pm$ 0.0089&0.7672 $\pm$ 0.0018\\
DeepAR&1.3185 $\pm$ 0.0826&3.0007 $\pm$ 0.0359&0.0355 $\pm$ 0.0008&2.1192 $\pm$ 0.0161&0.464 $\pm$ 0.0058\\
ETS&0.5137 $\pm$ 0.0758&0.2298 $\pm$ 0.0042&0.0332 $\pm$ 0.0004&1.2715 $\pm$ 0.0071& N/A \\
MQ-CNN&2.375 $\pm$ 0.1399&4.5208 $\pm$ 0.035&0.0481 $\pm$ 0.0062&3.4619 $\pm$ 0.4748&0.5667 $\pm$ 0.0096\\
NPTS&1.2909 $\pm$ 0.0622&2.5102 $\pm$ 0.0106&0.1235 $\pm$ 0.0002&1.7341 $\pm$ 0.0117&0.5871 $\pm$ 0.001\\
Rotbaum&0.3656 $\pm$ 0.0211&4.5134 $\pm$ 0.0157&0.0343 $\pm$ 0.0003&0.5403 $\pm$ 0.0032&0.438 $\pm$ 0.0015\\
\midrule
mean&0.5672 $\pm$ 0.0424&1.2186 $\pm$ 0.0046&0.0436 $\pm$ 0.0003&0.8564 $\pm$ 0.0438&0.4283 $\pm$ 0.0027\\
median&0.521 $\pm$ 0.0406&0.8816 $\pm$ 0.0057&0.0356 $\pm$ 0.0001&0.8325 $\pm$ 0.0159&0.374 $\pm$ 0.0023\\
GB ($D_1, D_{\text{BW}_1}$)&0.3044 $\pm$ 0.0346&0.2298 $\pm$ 0.0042&0.0329 $\pm$ 0.0001&0.5403 $\pm$ 0.0032&0.4054 $\pm$ 0.0051\\
Best ($D_1, D_{\text{BW}_1}$)&0.3656 $\pm$ 0.0211&0.2298 $\pm$ 0.0042&0.0332 $\pm$ 0.0004&0.5403 $\pm$ 0.0032&0.438 $\pm$ 0.0015\\
Unregularized&0.0589 $\pm$ 0.0009&0.2385 $\pm$ 0.0002&0.0274 $\pm$ 0.0001&0.134 $\pm$ 0.0007&0.343 $\pm$ 0.0007\\
Ours&\textbf{0.0587 $\pm$ 0.0007}& \textbf{0.1965 $\pm$ 0.0007}&\textbf{0.0268 $\pm$ 0.0003}&\textbf{0.1334 $\pm$ 0.0007}&\textbf{0.3294 $\pm$ 0.0016}\\
\bottomrule
\end{tabular}%
}

\vskip -0.1in
\end{table*}

\begin{table*}[h]
\centering
\caption{Mean weighted quantile losses over quantiles $\tau = [0.1, 0.5, 0.9]$ for various base learners and ensembling strategies on real-world open-source time series datasets, where Gaussian noise is added to the base learner predictions across items over 5 runs.}
\label{tab:syn_items}
\vskip 0.1in
\scalebox{0.75}{
\begin{tabular}{cccccc}
\toprule
Base Learner / \\ Ensemble Strategy & \texttt{Elec} &\texttt{Kaggle}&\texttt{M4-daily} &\texttt{Traf} &\texttt{Wiki}\\ 
\cmidrule(l){1-6}
ARIMA&0.2026 $\pm$ 0.0019&0.1223 $\pm$ 0.0001&0.0348 $\pm$ 0.0001&0.3707 $\pm$ 0.0012&0.6081 $\pm$ 0.0004\\
DeepAR&0.283 $\pm$ 0.011&0.2758 $\pm$  0.0051&0.0353 $\pm$ 0.0008&0.4442 $\pm$ 0.0053&0.3564 $\pm$ 0.0085\\
ETS&0.3534 $\pm$ 0.031&\textbf{0.1212  $\pm$ 0.0005}&0.0337 $\pm$ 0.0005&0.6846 $\pm$ 0.0027& N/A\\
MQ-CNN&0.3297 $\pm$ 0.0237&0.2752 $\pm$ 0.0009 &0.0425 $\pm$ 0.0033&0.4601 $\pm$ 0.0603&0.3796 $\pm$ 0.004\\
NPTS&0.314 $\pm$ 0.0151&0.1636 $\pm$  0.0001&0.1231 $\pm$ 0.0002&0.4158 $\pm$ 0.0023&0.4981 $\pm$ 0.0006\\
Rotbaum&0.437 $\pm$ 0.0269&0.2381 $\pm$ 0.0004&0.0355 $\pm$ 0.0003&0.4265 $\pm$ 0.0016&0.4649 $\pm$ 0.0039\\
\midrule
mean&0.1338 $\pm$ 0.0043&0.1593  $\pm$ 0.0005&0.0434 $\pm$ 0.0002&0.2218 $\pm$ 0.0034&0.3956 $\pm$ 0.0033\\
median&0.1526 $\pm$ 0.0065&0.1354  $\pm$ 0.0001&0.0357 $\pm$ 0.0002&0.2179 $\pm$ 0.0066&0.3539 $\pm$ 0.0031\\
GB ($D_1, D_{\text{BW}_1}$)&0.1333 $\pm$ 0.0039&0.1618  $\pm$ 0.001&0.0331 $\pm$ 0.0002&0.2099 $\pm$ 0.0053&0.3519 $\pm$ 0.0058\\
Best ($D_1, D_{\text{BW}_1}$)&0.2026 $\pm$ 0.0019&0.1223 $\pm$ 0.0001&0.0337 $\pm$ 0.0005&0.3707 $\pm$ 0.0012&0.3732 $\pm$ 0.0107\\
Unregularized&0.0521 $\pm$ 0.002&0.2209 $\pm$ 0.0006&0.0276 $\pm$ 0.0001&0.1125 $\pm$ 0.0003&0.3384 $\pm$ 0.0007\\
Ours&\textbf{0.0505 $\pm$ 0.0017}&0.156  $\pm$ 0.0008&\textbf{0.0268 $\pm$ 0.0003}&\textbf{0.1094 $\pm$ 0.0004}&\textbf{0.3235 $\pm$ 0.0016}\\
\bottomrule
\end{tabular}}

\end{table*}
\vspace*{-0.3cm}
\begin{table*}[h]
\centering
\caption{Mean weighted quantile losses over quantiles $\tau = [0.1, 0.5, 0.9]$ for various base learners and ensembling strategies on real-world open-source time series datasets, where Gaussian noise is added to the base learner predictions across quantiles over 5 runs.}
\label{tab:syn_quantile}
\vskip 0.1in
\scalebox{0.75}{
\begin{tabular}{cccccc}
\toprule
Base Learner / \\ Ensemble Strategy & \texttt{Elec} &\texttt{Kaggle}&\texttt{M4-daily} &\texttt{Traf} &\texttt{Wiki}\\ 
\cmidrule(l){1-6}
ARIMA&0.2347 $\pm$ 0.0067&0.1269 $\pm$ 0.0001&0.0344 $\pm$ 0.0&0.3877 $\pm$ 0.0008&0.606 $\pm$ 0.0005\\
DeepAR&0.789 $\pm$ 0.0573&0.3571 $\pm$ 0.0069&0.0369 $\pm$ 0.0007&1.2155 $\pm$ 0.013&0.4525 $\pm$ 0.004\\
ETS&0.4443 $\pm$ 0.0687&\textbf{0.1201 $\pm$ 0.0003}&0.0359 $\pm$ 0.0004&0.9363 $\pm$ 0.0111&N/A\\
MQ-CNN&0.9111 $\pm$ 0.0167&0.2904 $\pm$ 0.0013&0.0531 $\pm$ 0.0096&1.2778 $\pm$ 0.2005&0.4692 $\pm$ 0.0076\\
NPTS&0.6306 $\pm$ 0.024&0.2942 $\pm$ 0.0011&0.1269 $\pm$ 0.0002&0.9207 $\pm$ 0.0064&0.5888 $\pm$ 0.0017\\
Rotbaum&0.5671 $\pm$ 0.0525&0.2497 $\pm$ 0.0004&0.0374 $\pm$ 0.0005&0.71 $\pm$ 0.0024&0.5356 $\pm$ 0.0072\\
\midrule
mean&0.2511 $\pm$ 0.0173&0.1743 $\pm$ 0.0007&0.0441 $\pm$ 0.0004&0.3715 $\pm$ 0.0156&0.4107 $\pm$ 0.0032\\
median&0.2444 $\pm$ 0.0222&0.1394 $\pm$ 0.0003&0.036 $\pm$ 0.0002&0.3749 $\pm$ 0.0106&0.3664 $\pm$ 0.0027\\
GB ($D_1, D_{\text{BW}_1}$)&0.2268 $\pm$ 0.0103&0.1455 $\pm$ 0.0014&0.0335 $\pm$ 0.0002&0.3634 $\pm$ 0.0036&0.4057 $\pm$ 0.0046\\
Best ($D_1, D_{\text{BW}_1}$)&0.2347 $\pm$ 0.0067&0.1256 $\pm$ 0.0022&0.0344 $\pm$ 0.0&0.3877 $\pm$ 0.0008&0.4692 $\pm$ 0.0076\\
Unregularized&0.0555 $\pm$ 0.0006&0.235 $\pm$ 0.0003&\textbf{0.0268 $\pm$ 0.0001}&0.1233 $\pm$ 0.0005&0.3431 $\pm$ 0.0005\\
Ours&\textbf{0.0541 $\pm$ 0.0015}&0.1637 $\pm$ 0.0028&0.0269 $\pm$ 0.0002&\textbf{0.1228 $\pm$ 0.0004}&\textbf{0.3285 $\pm$ 0.001}\\
\bottomrule
\end{tabular}
}
\end{table*}

\vspace{-0.2cm}
\paragraph{Baselines.}
\label{sec:baselines}
We compare to the following simple ensemble baselines:
\begin{itemize}[noitemsep,topsep=0pt] 
    \item Mean: For each item, timestamp, and quantile, take a simple mean of all of the base learners.
    \item Median: For each item, timestamp, and quantile, take a simple median of all of the base learners.
    \item Global Best (GB) ($D_1, D_{\text{BW}_1}$): Of all of the possible combinations of algorithms to choose, choose the one for which the simple mean of the predictions of the chosen algorithms lead to the best performance on $ D_{\text{BW}_1}$. The same algorithms are chosen across timestamps, quantiles, and items. (Greedy ensemble, \cite{caruna2004greedy}, is an approximation of this method in the case that there are many base learners.)
    \item Best $ (D_1, D_{\text{BW}_1})$: Choose the single best base learner on $ D_{\text{BW}_1}$. 
    \item Unregularized: Use $\alpha=(0,0,0,0)$ (in the sense of the family given in Subsection \ref{sec:givefamily}); i.e., always use the unregularized stacked generalization, rather than choosing the best stacked generalization via cross validation. 
    
\end{itemize}

We remark that if we had used only one backtest window, namely if we learned $\alpha$ on the same backtest window on which the weights are optimized, then $\alpha$ would have to equal $(0,0,0,0)$, and so this setup is equivalent to this baseline.

\paragraph{Benchmarking with Raw Predictions.}

We first compare the predictions from our method to the raw predictions from the base learners, and the baseline ensembling methods. 

\makeatletter
\setlength{\@fptop}{0pt} 
\setlength\@fpsep{8pt plus 0.1fil} 
\makeatother

We remark that unlike in the literature about the ``forecast combination puzzle'' (Section \ref{related}), our method offers significant performance boost compared with simple average. We attribute this discrepancy to the idealized assumptions in the literature, most prominent among them being that they assume that the weights remain the same across timestamps or any other dimension. Table \ref{tab:regular} shows the results.

\paragraph{Synthetic Experiments Adding Noise to the Predictions Across Various Dimensions.}\label{synthsec}
We have seen in Table \ref{tab:syn_items} that our algorithm works well in a natural setting. But given that the family we have chosen (in Subsection \ref{sec:givefamily}) is meant to find the ideal degree to which the weights are forced to be similar across item/timestamps/quantiles, in this section we create synthetic experiments the stress-test this ability, and give strong evidence that it is able to adjust to these disruptions.

To be specific, for each set of base learner predictions $\eta_l(D_n)=\{\hat {z}^{(l)}_{i,j,k}(D_n)\}^{i,l}_{j,k}$, where $n=0,1,2$,  
  $i$ indexes the items, $j$  the forecast horizon, and $k$ the quantiles, we introduce Gaussian noise. Since we are applying the noise for each data training and validation splits, in the following we simply use the notation $\{\hat {z}^{( l)}_{i,j,k}\}^{i,l}_{j,k}$. Let $s^{(l)}_{i,k} = \text{std}_j(\{\hat z^{(l)}_{i,j,k}\})$ be the empirical standard deviation computed over the time dimension $j$ for fixed $i,k, l$. For each simulation, we let the selected noise be the same for all the backtest windows.  
For each $i,j,k$ and $l$, we then add Gaussian noise $\epsilon_{i,j,k}^{(l)}$ to the base learner prediction, i.e. \[\hat{z}^{(l)}_{i,j,k}\mapsto \hat{z}^{(l)}_{i,j,k} + N(0, \epsilon_{i,j,k}^{(l)}),\] where the noise $\epsilon_{i,j,k}^{(l)}$ varies in the following three ways: 
\begin{enumerate}[noitemsep,topsep=0pt] 
    \item Across time:  For each $l$, let $r_l \sim U(0,0.5)$ and $\epsilon_{i,j,k}^{(l)} = 2jr_ls^{(l)}_{i,k}/h$ 
    for forecast horizon $h$.
    In this experiment, we design the noise to be increasing with the time index $j$ to mimic the natural order of time, where the predictions become worse over time. (See Table \ref{tab:syn_time}).
    \item Across items: For each $l$ and $i=1,...,N$ let $r_{i,l} \sim U(0,2)$ and $\epsilon_{i,j,k}^{(l)} =\epsilon_{i,k}^{(l)} = r_{i,l} s^{(l)}_{i,k}$  (See Table \ref{tab:syn_items}).
    \item Across quantiles: For each $l$ and $k=1,2,3$, let $r_{k,l} \sim U(0,2)$ and $\epsilon_{i,j,k}^{(l)}=\epsilon_{i,k}^{(l)}=2r_{k,l}s^{(l)}_{i,k}$  
    (See Table \ref{tab:syn_quantile}).
\end{enumerate}

See Tables \ref{tab:syn_time} (adding noise across time) and \ref{tab:syn_items} (adding noise across items) and \ref{tab:syn_quantile} (adding noise across quantiles) for results. In these three synthetic experiments, we see that our method is successful in adjusting for the noises, and gives similar results in those in Table \ref{tab:regular}  with no noise added. We also see that using the unregularized loss, which corresponds to choosing $\alpha=(0,0,0,0)$ rather than learning it via cross-validation, is a surprisingly close second. It is not remarkable that the rest of the baselines do not perform well since the other baselines except for median use the same weights for all timestamps/quantiles/items, and median is another one size fits all solution that does not vary across timestamps/quantiles/items. 
\vspace{-0.2cm}
\section{Conclusion}
In this paper, we provide theoretical guarantees for families of stacked generalizations, which are commonly used in practice in state-of-the-art ensembling methods.  We extend the work in \citet{van2007super}  that shows that choosing the best stacked generalization from a finite family of constant stacked generalizations based on cross-validated performance does not perform much worse than the oracle best, both by allowing the stacked generalizations to be learned rather than constant, and by giving a result for finite-dimensional, rather than finite, families. We then show a particular family of stacked generalizations in the probabilistic time series forecasting use case. We formulate the problem of finding this family as a regularized regression problem, where the choice of regularizers parametrizes the family. Our experiments demonstrate that our intuition that designing a family of stacked generalizations in the time series case, where each member of the family gives different flexibility for the weights to vary across the various timestamps/quantiles/items is effective.

\clearpage

%\nocite{langley00}

\bibliography{ts.bib}

\bibliographystyle{icml2023}

\newpage
\appendix
\onecolumn

\section{Omitted Proofs}\label{omitted}
In this section we provide the proofs that were omitted from Section \ref{sec:theorem}.

Using the notation of Section \ref{prelim}, let $\mathbb{G}_i:=\sqrt{n_i}(\mathbb{P}_i-P)$ be the ``empirical process'' of $\mathbb{P}_i$ for $i=1,2$. The following lemma is a straightforward generalization of Lemma 2.1 in \cite{van2006oracle} to the case of infinitely many $\theta_{\alpha}$s and order-dependent algorithms.  As the proof is the same \it mutatis mutandis \rm as in \cite{van2006oracle}, we omit it.

\begin{lemma}\label{lemlem}
For every $\hat{\alpha}\in \mathcal{J}$, let 
\begin{equation*}
\begin{aligned}
W_{\hat{\alpha}}:=\Big\{\tilde{\alpha}\in \mathcal{J}\Big|\int L_{\hat \alpha}(D_0)(X)d\mathbb{P}_1\\\quad\leq \int L_{\tilde \alpha}(D_0)(X)d\mathbb{P}_1\Big\},    
\end{aligned}
\end{equation*}
be the set of indices $\tilde{\alpha}$ for which $\theta_{\hat{\alpha}}$ outperforms $\theta_{\tilde{\alpha}}$ on validation. Then for any $\delta\geq -1$:
\begin{align}
\begin{aligned}
  &\EE(\int L_{\hat \alpha}(D_0)(X)dP)\leq\\&\quad(1+2\delta)\inf_{\tilde{\alpha}\in W_{\hat{\alpha}}}\left(\EE(\int L_{\tilde \alpha}(D_0)(X)dP)\right) \\
&\quad{+}\:\frac{1}{\sqrt{n_1}}\EE(\sup_\alpha\int L_{\alpha}(D_0)(X)d((1+\delta)\mathbb{G}_1-\delta\sqrt{n_1} P)) \\
&\quad{+}\:\frac{1}{\sqrt{n_1}}\EE(\sup_\alpha -\!\!\!\int L_{\alpha}(D_0)(X)d((1+\delta)\mathbb{G}_1+\delta \sqrt{n_1}P)).
\end{aligned}
\end{align}
\end{lemma}

Our goal, therefore, is to bound the last two terms. In the case that $|\mathcal{J}|$ is finite this was done in Lemma 2.2 of \cite{van2006oracle}.

\begin{lemma}\label{bound} (Lemma 2.2 of \cite{van2006oracle})
Assume that $\mathcal{J}$ is finite, that $\mathcal{F}$ has Bernstein numbers $(M(\mathcal{F}),v(\mathcal{F}))$, and that the $\theta_{\alpha}$'s are invariant of the training dataset's order. Then for every $\delta> 0$ and $1\leq p\leq 2$:
\begin{equation}
\begin{aligned}
&\EE(\max_{f\in\mathcal{F}}\pm\int fd(\mathbb{G}_1-\delta\sqrt{n_1}P))\leq\\&\quad\frac{8}{n_1^{\frac1p-\frac12}}\log(1+|\mathcal{J}|)\max_{f\in\mathcal{F}}\left(\frac{M(\mathcal{F})}{n_1^{1-\frac1p}} + \left(\frac{v(\mathcal{F})}{(\delta\int fdP)^{2-p}}\right)^{\frac1p}\right).    
\end{aligned}
\end{equation}
\end{lemma}

While the lemma assumes the $\theta_{\alpha}$'s are invariant of the training dataset order, this assumption is in fact not used in its proof. Thus it suffices to reduce the case that $\mathcal{J}$ is finite dimensional to the lemma above. This is done via a discretization of $\mathcal{J}$ itself. (Note that this is not the same as discretizing the set of functions that the stacked generalizations may become.) 

\begin{theorem}\label{generalized}
Under the assumptions of Theorem \ref{mainthm}, for every $\delta> 0$, $1\leq p\leq 2$, and sequence $\varepsilon_{n_1}>0$:

\begin{align}
\begin{aligned}
&\EE(\sup_{f\in\mathcal{F}}\pm\int fd(\mathbb{G}_1-\delta\sqrt{n_1}P))\leq\\&\quad\sup_{f\in\mathcal{F}}\left(\frac{8\left(\frac{M(\mathcal{F})}{n_1^{1-\frac1p}} + \left(\frac{v(\mathcal{F})}{(\delta\int fdP)^{2-p}}\right)^{\frac1p}\right)\log(1+N^{\text{int}}(\mathcal{J},\frac{\varepsilon_{n_1}}{\ell}))}{n_1^{\frac1p-\frac12}}\right)\\
&\quad{+}\:((1+\delta)\sqrt{n_1}+\frac{1}{\sqrt{n_1}})\varepsilon_{n_1}.
\end{aligned}
\end{align}
\end{theorem}
\vspace{-.5cm}

\begin{proof} (of Theorem \ref{generalized})\,\\
First note that if  $\|f-f' \|_{\infty}<\varepsilon_{n_1}$ then 
\[
\begin{aligned}
|\mathbb{G}_1f-\delta\sqrt{n_1}Pf-(\mathbb{G}_1f'-\delta\sqrt{n_1}Pf')| &\leq
|\mathbb{G}_1f-\mathbb{G}_1f'| + \delta\sqrt{n_1}|P(f-f')| \\
&\leq
n_1^{-\frac12}\sum_{l=1}^{n_1} |f(X_l^1)-f'(X_l^1)|+(n_1^{-\frac12}+\delta\sqrt{n_1})|P(f-f')|\\
&\leq \sqrt{n_1}\varepsilon_{n_1}+(n_1^{-\frac12}+\delta\sqrt{n_1})\varepsilon_{n_1} \\
&=((1+\delta)\sqrt{n_1}+\frac{1}{\sqrt{n_1}})\varepsilon_{n_1},
\end{aligned}
\]
and therefore, after taking suprema and expected values, we get that for every subset $\mathcal{F}'$ of $\mathcal{F}$ for which every element in $\mathcal{F}$ is $\varepsilon_{n_1}$-close to an element in $\mathcal{F}'$
$$\EE(\sup_{f\in\mathcal{F}}\pm\int fd(\mathbb{G}_1-\delta\sqrt{n_1}P))\leq \EE(\sup_{f'\in\mathcal{F}'}\pm\int f' d(\mathbb{G}_1-\delta\sqrt{n_1}P)) + ((1+\delta)\sqrt{n_1}+\frac{1}{\sqrt{n_1}})\varepsilon_{n_1}.$$

We now proceed to reduce Theorem \ref{generalized} to Lemma \ref{bound} above, which applies to the case of finite $\mathcal{J}$. By our assumptions, we have that $N^{\text{int}}(\mathcal{F}, \ell\varepsilon)\leq N^{\text{int}}(\mathcal{J},\varepsilon)$, or in other words $N^{\text{int}}(\mathcal{F}, \varepsilon)\leq N^{\text{int}}(\mathcal{J}, \frac{\varepsilon}{\ell})$. 

Combining Lemma \ref{bound} applied to the centers $\mathcal{F}'$ of a minimal $\varepsilon_{n_1}$-covering of $\mathcal{F}$ with the inequality above, the result follows.
\end{proof}
\begin{proof} (of Theorem \ref{mainthm})\,\\
The main portion of the theorem follows directly from Theorem \ref{generalized} and Lemma \ref{lemlem} by bounding each of the last two terms in Lemma \ref{lemlem} using Theorem \ref{generalized} and then multiplying by the prefactor $1/\sqrt{n_1}$ (and summing the two contributions).

The second portion of the theorem follows from the inequality $N^{\text{int}}(\mathcal{J}, \frac{\varepsilon}{\ell})\leq N^{\text{ext}}(\mathcal{J}, \frac{\varepsilon}{2\ell})$ (with $N^{\text{ext}}$ representing the external covering number), as well as the fact that if $\mathcal{J}$ is contained in a $K$-ball around the origin in an ambient Euclidean space of dimension $M$, then we have the following inequality for the external covering number:$$N^{\text{ext}}\bigg(\mathcal{J},\frac{\varepsilon_{n_1}}{2\ell}\bigg)\leq \left(\frac{4\ell K\sqrt{M}}{\varepsilon_{n_1}}\right)^M.$$ Therefore if one chooses $\varepsilon_{n_1}:=n_1^{-\frac12-\epsilon}$ for some $\epsilon>0$, then $\log(1+N^{\text{int}}(\mathcal{J},\frac{\varepsilon_{n_1}}{2\ell}))$ is order of magnitude of $\log(n_1)$. Further, if one wants to include $p$ additional algorithms as stacked generalizations then that can be done by adding them into both $\mathcal{F}$ and $\mathcal{F}'$ in the proof of Theorem \ref{generalized}, and therefore results in the log term described.
\end{proof}
\clearpage

\section{Tabular Analogue of the Family in Section \ref{timefamily}}
In this appendix we point out that the construction in Section \ref{timefamily} can be employed in the tabular situation as well, for any case where we have a decomposition of the prediction space $\mathbb{R}^{r'}\cong\mathbb{R}^{r_1'}\bigotimes\cdots\bigotimes\mathbb{R}^{r_p'}$ of the base learners $\eta_1,...,\eta_m$ into a tensorial factors. For $\alpha:=(\alpha_1,...,\alpha_p)$ we construct a stacked generalization $A_{\alpha}(D_0)$ as $(x, \eta_1(D_0)(x),...,\eta_m(D_0)(x))\mapsto\sum_{\forall j\,\,1\leq d_j \leq r_i'}\left(\sum_l w_{(d_i)_{i=1}^p,l}\eta_l(D_0)(x)\right)e_{d_1}\otimes\cdots\otimes e_{d_p}$. The weights are chosen to minimize
\begin{equation}
f_{\alpha}(D_0):=\sum_{X\in D_0} L_{\alpha}(X)+\sum_{i=1}^p\alpha_i\mathbb{H}(\sigma^{(i)}(w))
\label{eq:obj_tab}
\end{equation}
where $\mathbb{H}^{(i)}$ is the entropy term similar to Section \ref{timefamily}. It is also possible to add other regularization terms such as $l_1$ regularization as we did in Section \ref{timefamily}.

\clearpage

\section{Experimental Detail}
\subsection{Probabilistic Time Series Forecasting Problem Definition}
\label{sec:pts_prob_def}
We first define the time-series data as $D := \{z_{i,j'}\}^i_{j'}$, $z_{i,j'}\in\mathbb{R}$,
 such that $i = 1,2,\dots,N$ denotes the $i^{th}$ time-series (henceforth, item) with a total of $N$ items, and for each item-$i$, $j'=1,2,\dots,T_i$ 
 denotes the time-points with the historical length $T_i$. For simplicity, we assume $\forall i\,\,T_i = T$.
The goal of a probabilistic time-series forecasting algorithm is to output a probabilistic prediction for the future $h$ time-steps for each time-series, where $h$ is pre-determined, for quantiles $0\leq\tau_1,\dots,\tau_q\leq 1$. For the notational purpose and to avoid repetition, we reserve $i = 1, 2, \dots, N$, and $j = 1, 2, \dots, h$ 
, and $k = 1, 2, \dots, q$ through rest of the paper, and henceforth drop the index range for $i,j,k$. The output consists of estimates $\{\hat{z}_{i,j,k}\}^{i}_{j,k}$ of the $\tau_k$-quantiles of $z_{i, T+j}$ conditioned on historical data.

Given computed predictions $\hat{Z}(D):=\{\hat{z}_{i,j,k}\}^{i}_{j,k}$ and the true values 
$D_{\text{BW}} := \{z_{i,T+j}\}^{i}_{j},$ 
we aim to minimize the mean weighted quantile loss:
\begin{equation}\label{eq:mean_QL}
\begin{aligned}
& L(\hat{Z}(D), D_{\text{BW}}):=
\frac{2}{q}\frac{\sum\nolimits_{i,j,k} \max\{\tau_k(z_{i, T+j}-\hat{z}_{i,j,k}), (1-\tau_k)(\hat{z}_{i,j,k}-z_{i, T+j})\}}{\sum\nolimits_{i,j}\vert z_{i,T+j}\vert}.
\end{aligned}
\end{equation}

\subsection{Family of Stacked Generalizations for Probabilistic Time Series Forecastings}
\label{app:method}
Let $\eta_1,\dots,\eta_m$ be $m$ arbitrary probabilistic time-series forecasting algorithms. In order to simplify the notation, if $D$ is time-series data as defined in Section\,\ref{sec:pts_prob_def}, we let $\eta_l(D)$ be $\eta_l$ trained on $D$ and then making inferences on $D$; i.e., $\eta_l(D) = \{\hat{z}^{(l)}_{i,j,k}(D)\}_{j,k}^i$, which consists of the predictions for each item in $D$ for the $h$ many unseen timestamps into the forecast horizon after the last seen value for $q$ quantiles; i.e., $\eta_l(D) = \{\hat{z}^{(l)}_{i,j,k}(D)\}_{j,k}^i$.

Since cross-validation is more subtle in the time series use case, we introduce some additional notation.  
In particular, we define training and backtest window datasets pairs, where a backtest window consists of the true values on a subsequent window of length $h$ after the last training time point as: 
\begin{equation}
    \begin{aligned}
        (D_n, D_{\text{BW}_n}) &:= ( 
        \{z_{i,j'}\}_{j'}^i,
        \hspace{0.1cm}
        \{z_{i,T-\mu h+j}\}_j^i),
        \label{eq:train_val}
    \end{aligned}
\end{equation}
for $j' = 1, \dots, T - \mu h$, $\mu=2-n$, $n = 0, 1, 2$. 
Here we use two pairs of training and validation sets, that is ($D_0, D_{\text{BW}_0}$) and ($D_1, D_{\text{BW}_1}$). Note that $D$ and $D_{\text{BW}}$ in Section \ref{sec:pts_prob_def} are equivalent to to $D_2$ and $D_{\text{BW}_2}$, respectively. 

For every choice of ensemble weights $w = \{w^{(l)}_{i,j,k}\}_{j,k}^{i, l}$, $l=1,2,\dots,m$ we denote: 

\begin{equation}
\hat{Z}(D, w) := 
\left\{ \sum_l w^{(l)}_{i,j,k}(D)\hat z^{(l)}_{i,j,k}(D)\right\}^i_{j,k},
 \label{eqn:stacked_gen_ts}
 \end{equation}

as the weighted ensemble combination. The weights are restricted to sum up to $1$; namely, for every $i,j,k$ we have that $\sum_l w^{(l)}_{i,j,k}(D)= 1$. We let 
\begin{equation}\label{eqn:obj_ts}
\begin{aligned}
&f_{\alpha}[(D_n, D_{\text{BW}_n}), w] = L\big(\hat{Z}(D_n, w), D_{\text{BW}_n}\big)
+\sum_{d=1}^3\alpha_d\mathbb{H}(\sigma^{(d)}(w))
 + \alpha_{4}\sum\limits_{i,j,k,l}|w^{(l)}_{i,j,k}|,
\end{aligned}
\end{equation}
be the regularized mean weighted quantile loss with $L$ given by Eqn. \eqref{eq:mean_QL} and ensemble estimate $\hat{Z}(D_n, w)$ by Eqn. \eqref{eqn:stacked_gen_ts}.  We use an entropy of softmax regularizer: 
$$
\mathbb{H}(\sigma^{(d)}(w)) := \sum\limits_{i,j,k,l}\sigma^{(d)}(w^{(l)}_{i, j, k})\log(\sigma^{(d)}(w^{(l)}_{i, j, k})),
$$ where $\sigma^{(d)}(w^{(l)}_{i, j, k}) = \exp(w_{i,j,k}^{(l)}) / \sum_{\mathcal{I}(d)} \exp(w_{i,j,k}^{(l)})$ denotes the softmax with the denominator summed over the double summations specified by $d$, that is, $\mathcal{I}(1), \mathcal{I}(2), \mathcal{I}(3)$ are $(j,k), (i,k), (i,j)$ respectively for $d=1,2,3$. 
The role of the entropy regularizing parameters  $\{\alpha_d\}_d$ for $d = 1, 2, 3$ in this family is to control how much the ensemble weights are allowed to vary across items, timestamps in the forecast horizon, quantiles, respectively. Lastly, $\alpha_4$ denotes the $l_1$ regularizer parameter, which specifies how much to be biased towards preferring a single algorithm versus diversifying. 

We follow Algorithm \ref{alg:compute_weights} to compute the optimal ensembling weights $w^*,$ and output the optimal $\tilde \alpha$ ensembling member from our family of stacked generalizations:
\begin{equation}
    A_{\tilde{\alpha}}(D_2):=\hat{Z}(D_2,w^*),
    \label{eqn:final_ensemble}
\end{equation}
where $\hat{Z}$ is the weighted linear combination of base learners in Eqn. \eqref{eqn:stacked_gen_ts}.
 \clearpage
\section{Towards a Uniform Treatment of Both Tabular and Time Series Data}\label{samenotation}
A prominent difference in the way that we have treated tabular and time series data is in the meaning of ``cross validation''. In Section \ref{prelim}, the validation set consists of a different set of datapoints, whereas in Section \ref{timefamily} the validation set is in the same time series but with a different forecast horizon. In this appendix we will attempt to unify the notation for both cases.

The way to bridge this gap is to replace the dimension $d$ of feature space, from Section \ref{prelim}, with the first infinite ordinal. In more colloquial terms, feature space will be the disjoint union $\bigcupdot_{d\in\mathbb{N}}\mathbb{R}^d$. We keep the dimension of prediction space fixed, since in our time series forecasting setup infernces are always fixed size (length of forecast horizon times number of items times number of quantiles). In this way we may phrase cross validation for time series in the terminology of Section \ref{prelim}, by allowing each dataset to consist of a single datapoint. To wit, let $D_n, D_{\text{BW}_n}$ for $n=0,1$ be as in Section \ref{timefamily}, and let $\tilde{D}_n:=\{(D_n,D_{\text{BW}_n})\}$ for $n=0,1$. Then cross validation in the sense of Section \ref{timefamily} is simply cross-validation in the sense of Section \ref{prelim} for $\tilde{D}_0$ and $\tilde{D}_1$.

In order to apply any of the theory of the tabular case, we must endow $(\bigcupdot_{d\in\mathbb{N}}\mathbb{R}^d)\times \mathbb{R}^{r'}$, where $r'$ is the dimension of prediction space, with a topology and a probability measure. While this space has a natural topology (which is even metrizable) by letting each $\mathbb{R}^d$ be a separate connected component in $\bigcupdot_{d\in\mathbb{N}}\mathbb{R}^d$, this topology does not jibe with intuition. For example, two time series that are identical except that one has one additional timestamp compared to the other would be very far away. Perhaps a more natural choice would be a topology induced by dynamic time warping. Once a topology is fixed, one has to choose a (Borel) probability measure on $(\bigcupdot_{d\in\mathbb{N}}\mathbb{R}^d)\times \mathbb{R}^{r'}$. 

In whatever way that these choices are done, it is unrealistic to assume that the datapoint in $\tilde{D}_0$ is independent of the one in $\tilde{D}_1$. This implies that cross validation in the time series case cannot truly be brought into the fold of the tabular theory. Nevertheless, the tabular theory is directional.

\clearpage

\section{Dataset Details}\label{detailsappendix}
Table \ref{tab:datasets} lists the details of the datasets that we use in our experiments.
\begin{table}[h]
\centering
\caption{Summary of dataset statistics, where \texttt{Elec} and \texttt{Traf} are from the UCI data repository (\cite{Dua:2017}), \texttt{Kaggle} and \texttt{Wiki} from Kaggle (\cite{lai_dataset_2017}), and 6 different \texttt{M4} competition datasets (\cite{makridakisM4concl}).}
\label{tab:datasets}
\vskip 0.1in
\resizebox{0.7\linewidth}{!}{
\begin{tabular}{|l|l|c|c|c|c|c|c|c|c}
\hline
\textsc{domain} & \textsc{name} & \textsc{support} & \textsc{freq} & \textsc{no. ts} & \textsc{avg. len} &  \textsc{pred. len}

\\
\hline
 electrical load & \texttt{Elec} & $\mathbb{R}^+$ & H & 321 & 21044 & 24 \\
 \hline 
Rossman  & \texttt{Kaggle}  &$\mathbb{N}$ & D & 500 & 1736 & 90\\

  \hline
  \begin{tabular}[c]{@{}l@{}}M4 forecasting \\ competition \\   \end{tabular} & \texttt{M4-daily} & $\mathbb{R}^+$ & D & 4227  & 2357 &14 \\
  \hline
road traffic & \texttt{Traf} & $[0, 1]$ & H & 862 & 14036 & 24\\
\hline
\begin{tabular}[c]{@{}l@{}}visit counts of \\ wikipedia pages \\  \end{tabular} &
    \texttt{Wiki}    & $\mathbb{N}$ & D   & 9535                                                      & 762  & 7 \\
    \hline
\end{tabular}}
\end{table}

\label{rest}

\end{document}